\documentclass[preprint,12pt, a4paper]{elsarticle}

\usepackage{amssymb}
\usepackage{amsfonts}
\usepackage{adjustbox}
\usepackage{amsthm}
\usepackage{mathtools}
\usepackage{lineno}
\usepackage{hyperref}
\usepackage{float}
\usepackage{wrapfig}
\graphicspath{ {fig/} }
\usepackage{dsfont}
\restylefloat{table}
\usepackage[table,dvipsnames]{xcolor}

\usepackage{geometry}
\journal{SoftwareX}

\begin{document}

\begin{frontmatter}



\title{\textbf{tsrobprep - an R package for robust preprocessing of time series data}}


\author{Micha{\l} Narajewski}
\author{Jens Kley-Holsteg}
\author{Florian Ziel}

\address{University of Duisburg-Essen}

\begin{abstract}
Data cleaning is a crucial part of every data analysis exercise. Yet, the currently available \texttt{R} packages do not provide fast and robust methods for cleaning and preparation of time series data. The open source package \texttt{tsrobprep} introduces efficient methods for handling missing values and outliers using model based approaches. For data imputation a probabilistic replacement model is proposed, which may consist of autoregressive components and external inputs. For outlier detection a clustering algorithm based on finite mixture modelling is introduced, which considers time series properties in terms of the gradient and the underlying seasonality as features. The procedure allows to return a probability for each observation being outlying data as well as a specific cause for an outlier assignment in terms of the provided feature space. The methods work robust and are fully tunable. Moreover, by providing the \texttt{auto\_data\_cleaning} function the data preprocessing can be carried out in one cast, without comprehensive tuning and providing suitable results. The primary motivation of the package is the preprocessing of energy system data. We present application
for electricity load, wind and solar power data.

\end{abstract}


\end{frontmatter}

\section*{Current code version}

\begin{table}[H]
\begin{tabular}{|l|p{6.5cm}|p{6.5cm}|}
\hline
\textbf{Nr.} & \textbf{Code metadata description} & \textbf{Please fill in this column} \\
\hline
C1 & Current code version & 0.3.1 \\
\hline
C2 & Permanent link to code/repository used for this code version & \url{https://cran.r-project.org/web/packages/tsrobprep} \\
\hline
C3 & Legal Code License   & MIT license (MIT) \\
\hline
C4 & Code versioning system used & git \\
\hline
C5 & Software code languages, tools, and services used & R \\
\hline
C6 & Compilation requirements, operating environments \& dependencies & R ($\ge$ 3.2.0) \\
\hline
C7 & If available Link to developer documentation/manual & \url{https://cran.r-project.org/web/packages/tsrobprep/tsrobprep.pdf} \\
\hline
C8 & Support email for questions & michal.narajewski@uni-due.de \\
\hline
\end{tabular}
\end{table}


\section{Introduction}
The increasing availability of high frequency data measured, recorded and processed in real-time leverage the potential benefit of data analytics in day to day business. However, as the data is usually measured by sensors and real measurement stations, measurement and transmission failures are inevitable so that a data preprocessing is required. Focusing on energy systems, two aspects are crucial, firstly, the accrued data is often immediately processed so that fast and automatic algorithms are needed. Secondly, the data is usually unlabelled so that the detection requires purely data-driven approaches. Hence, the data cleaning and preprocessing requires an automatic, robust and mainly data-driven procedure, which is easily tunable and applicable in real-time to provide verified data for a safe operation of the energy system.

In the literature several R packages are available, which address the task of data cleaning and preparation. On the one hand, there are several packages specifically designed for dealing with missing values \cite{moritz2015}. Here, the packages are mainly focused either on data imputation in multivariate data by utilizing inter-attribute correlations as done in Amelia \cite{Honaker2011}, mice \cite{Buuren2011} and VIM \cite{Kowarik2016}, or on univariate data by employing time dependencies as done in imputeTS \cite{Moritz2017}, zoo \cite{Zeileis2005} and forecast \cite{Hyndman2008}. On the other hand, there are several packages specifically designed for dealing with outlying data. Here, among others HDoutliers \cite{Fraley2020}, mvoutlier \cite{Filzmoser2018}, and tsoutliers \cite{Lacalle2019} can be named. However, packages combining both outlier detection and data imputation methods for time series data in a robust and efficient way, as implemented e.g. in forecast \cite{Hyndman2008} and imputeFin \cite{Liu2021} for univariate data, are rare.  

The presented package tsrobprep is specialized on dealing with outliers and missing values in time series data in a robust and simple applicable way. It offers as core functions an algorithm for modelling missing values and an algorithm for detecting outliers. To enhance usability the package comes with three additional side functions, a robust decomposition algorithm, an algorithm to impute the modelled values, and a wrapper function to apply the former named functions at once. The first core function for modelling missing values is based on a two-step approach, which allows to deal with autoregressive effects and given external inputs. The second core function for outlier detection is based on a feature based clustering approach with finite (Gaussian) mixture models. Both core functions provide a probabilistic framework such that for data imputation a whole distribution can be modelled and for outlier detection the probability being an outlying data point can be provided. Focused on robustness, efficiency and usability the package tsrobprep is perfectly suited for cleaning various time series data sets in a univariate and multivariate setting, as done e.g. in \cite{Narajewski2020}. The package is available online in the CRAN repository.

The paper is structured as follows. Section 2 describes the utilized methods in detail and Section 3 presents the usability and data preprocessing potential. Section 4 compares the methods to competitors from literature. Finally, Section 5 discusses the impact of the package and concludes the paper.

%
%
%
%

\section{Software description}
For dealing with missing values, the \verb+model_missing_data+ function is proposed for handling univariate and multivariate time series data. It takes the time series data to be cleaned as an argument and returns the modelled missing values to be imputed. The function uses a two-step model-based approach to replace missing values. First, the data is decomposed to trend, seasonal and external input components, as well as to their interactions. This is done using the \verb+robust_decompose+ function that is available in the package. Then, on the remainder we utilize the following regression
\begin{equation}\label{eq:model_missing}
	\text{Target}_{\tau}(Y_t) = \beta_{0,\tau} + \sum_{p \in \mathcal{L}} \beta_{p,\tau} Y_{t-p} + \sum_{i \in \mathcal{I}} \beta_{i,\tau} X_{i,t} + \varepsilon_t = \boldsymbol{\beta}_{\tau} \mathbb{X}_t + \varepsilon_t,
\end{equation}
where $\text{Target}_{\tau}$ is either the expected value or a $\tau$-quantile, $Y_t$ is the time series remainder with the missing data to be replaced, and $X_{i,t}$ are the decomposition components (including external regressors) to allow in a multivariate framework for utilizing beside time-dependencies also inter-attribute correlations. Additionally, $\mathcal{L}$ is a set of lags for the autoregression, and $\mathcal{I}$ is a set of the decomposition components. By allowing for positive and negative lags in $\mathcal{L}$, past and forthcoming (if available) values can be utilized for modelling.
By default, the set of lags $\mathcal{L}$ takes reasonable values, depending on the provided vector of seasonalities $\boldsymbol{S}=(S_1,\ldots,S_K)$ and the input time series data. 
Precisely,
$\mathcal{L}$ contains all lags except those which involve multiple seasonal parameters and their negative values of a multi-seasonal AR($p+1,p,\ldots,p$) with seasonalities $\boldsymbol{S}$ and $p$ depending on data size.
For example for $\boldsymbol{S}=(24,168)$ and $p=1$, this is
$\mathcal{L} = \{1,2,24,25,168,169, -1,-2,-24,-25,-168,-169\}$.

If the user provides a matrix of external inputs, the algorithm allows to select the most appropriate inputs by user-tunable thresholds based on the linear correlation between the externals and the explained data. To utilize not only the inter-attribute information and autoregressive effects also lags for the externals can be applied for modelling. In case of the Target being the expected value the weighted lasso implemented in function \verb+glmnet+ in the \verb+glmnet+ package \cite{friedman2010glmnet} can be applied. Otherwise, a quantile regression implemented with the fast Frisch-Newton algorithm in function \verb+rq.fit.fnb+ in \verb+quantreg+ package~\cite{quantreg} can be applied. Moreover, the algorithm uses recursive replacement by default, which improves the modelling accuracy in the case of larger gaps, but it may also cause an error accumulation, so we give the user the option to switch it off.

For dealing with outlying data, the second core function \verb+detect_outliers+ is proposed for handling univariate data. It takes the time series data to be cleansed and returns the position of the detected outliers, the probability of being an outlying instance, the applied feature matrix and the specific cause for an outlier assignment in terms of the applied features. By recursive application within the wrapper function \verb+auto_data_cleaning+ also multivariate data sets can be handled. The algorithm can be tuned to control robustness, computational time and detection sensitivity. By utilizing specifically designed features for time series data, various patterns can be identified. As features, the gradient, absolute gradient, relative gradient, deterministic seasonal trend, seasonal gradient and absolute seasonal gradient are applied. For example, the absolute gradient tracks two-sided jumps with regard to the closest neighbours, the absolute seasonal gradient works the same, however, applied to the closest neighbours in terms of the provided seasonality $\boldsymbol{S}$. The relative gradient in turn tracks two sided jumps in relation to the local seasonal variance to account for moderate spikes in low volatility periods. By applying an unsupervised learning approach the procedure is well suited to handle unlabelled and unstructured data. The approach is based on model-based clustering with finite mixture models as outlined in \cite{Fraley2002}. It is assumed that the multivariate distribution of the feature space is a mixture of $G$ components, the likelihood for a mixture model can be defined as

\begin{equation}\label{eq:detect_outliers}
  \boldsymbol\ell_{MIX}(\theta_1,...,\theta_G | \textbf{y}) = \prod_{i=1}^{n} \sum_{k=1}^{G} \pi_k f_k(\textbf{y} | \theta_k),
\end{equation}

where \textbf{y} is the observation vector $y_1,...,y_n$, $f_k$ is the density (assumed to be multivariate normal (Gaussian)), $\theta_k$ are the parameters of the $k$-th component in the mixture and $\pi_k$ is the probability that an observation belongs to the $k$-th component. Note that $\sum_{k=1}^{G} \pi_k = 1$. 
Within the mixture one component is intended to represent outlying data. This component is introduced by artificially augmenting the number of outliers in the data set by blowing up the covariance matrix of the mixture. For modelling the data as a Gaussian finite mixture with different covariance structures and numbers of mixture components the \verb+Mclust+ function in \verb+mclust+ package \cite{Scrucca2016} is applied. For estimation the expectation-maximization algorithm initialized by hierarchical model-based agglomerative clustering is used as outlined in \cite{Fraley2002}. To speed up the computational time and robustness of the method, the estimation is carried out repeatedly on varying data subsets. To provide the specific cause for an outlier assignment, each observation, which has been identified as an outlying data point is shifted to the corresponding feature and multivariate feature combination mean, respectively. Then, the probability of being outlying data is recomputed allowing to name a specific feature or a combination of features, which caused the outlier assignment. 

\begin{figure}[b!]
	\includegraphics[width = 1\linewidth]{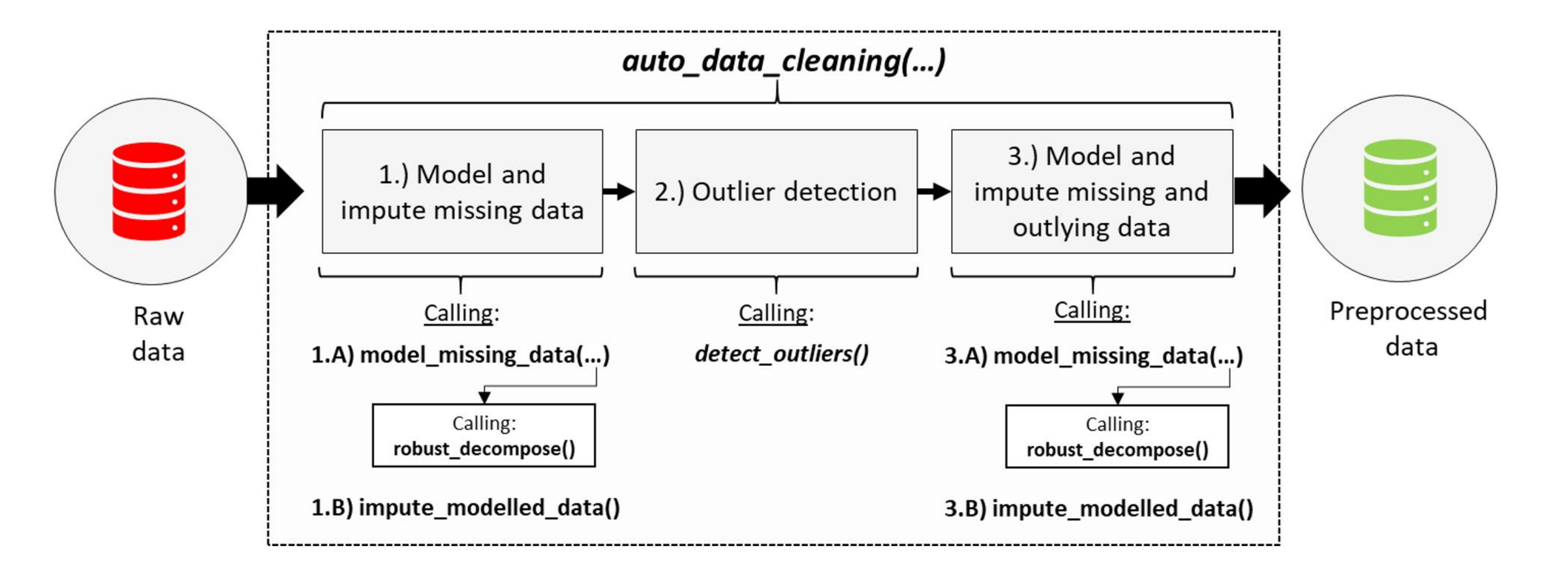}
	\caption{Workflow of the auto data cleaning function.}
	\label{fig:auto_cleaning_chart}
\end{figure}

The auxiliary function \verb+impute_modelled_data+ eases the usage of the package. It takes as argument a \texttt{tsrobprep} object, i.e. the object returned by the \verb+model_missing_data+ function, imputes the replaced values for given quantiles $\tau$ and returns a list of matrices (each element of the list corresponds to each quantile). The \verb+auto_data_cleaning+ function combines everything described above. First, it replaces missing values (if there are any) with a median forecast. This step is required as the outlier detection algorithm does not allow for missing values. Second, unreliable outliers are detected so that in the third step, the detected outliers and the initially modelled missing values are modelled and imputed in one cast. The repetitive modelling of the initially modelled missing values is reasonable as potential outliers might have negatively influenced the modelling process in the first run. The workflow of the function is presented in Figure~\ref{fig:auto_cleaning_chart}.

\section{Illustrative Examples}

\begin{figure}[b!]
	
	\includegraphics[width=1\linewidth]{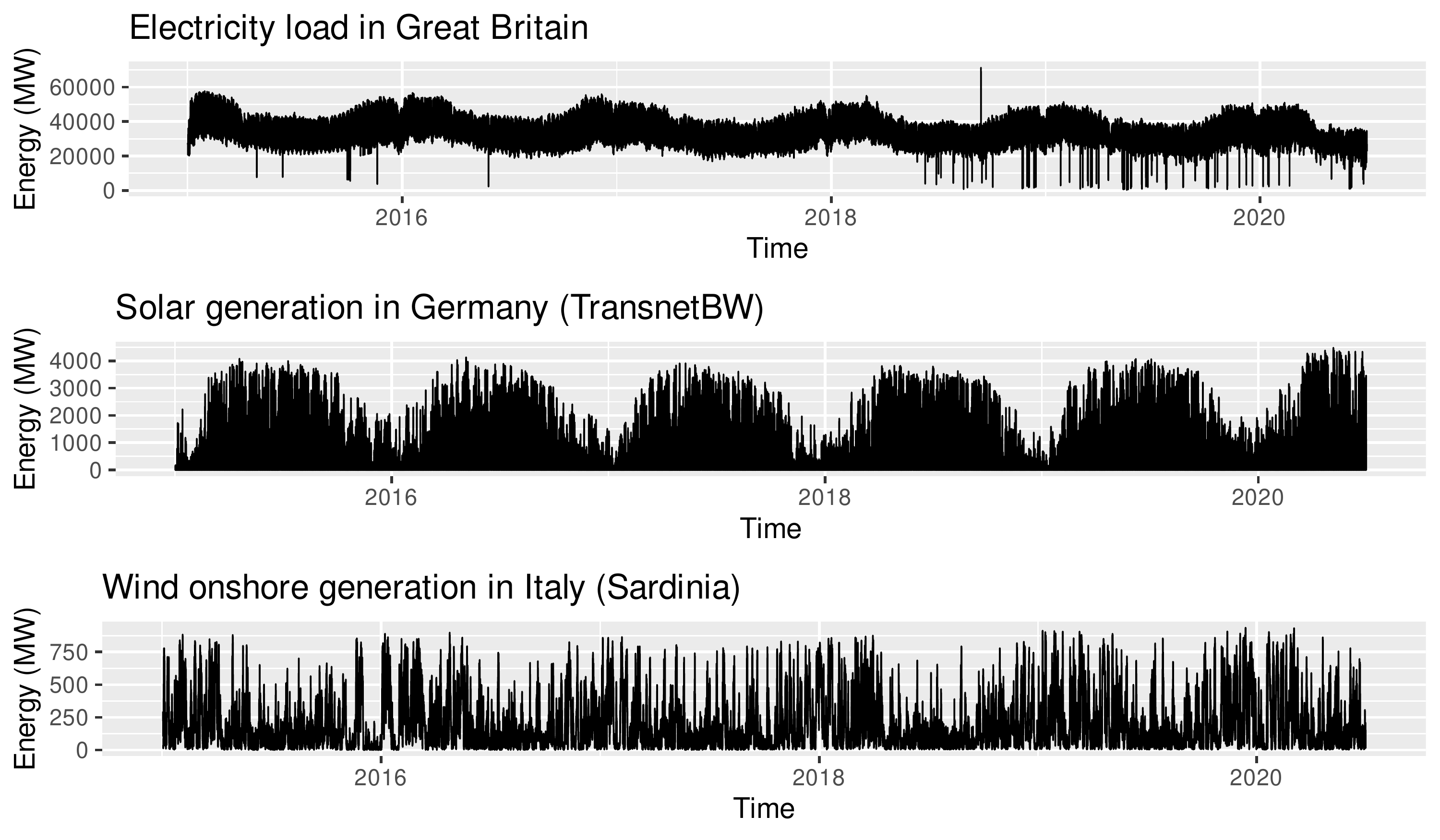}
	
	\caption{Three exemplary data sets containing both missing values and unreliable outliers.}
	\label{fig:initial_plots}
\end{figure}

We present the potential of the package by using its core functions on three time series data sets: the electricity load in Great Britain, the solar generation in Germany (TransnetBW control zone), and the wind onshore generation in Italy (Sardinian control zone). All these data sets span the date range from January 01, 2015 to June 30, 2020. The first time series is half-hourly, i.e. contains 48 values per day, the second quarter-hourly, i.e. contains 96 values per day, and the third is hourly, i.e. contains 24 values per day. The utilized data is freely available at ENTSOE Transparency platform \cite{entsoe}.
The considered time series are plotted over time in Figure~\ref{fig:initial_plots}. We can clearly see that the time series of electricity load in the United Kingdom exhibits many outliers that are very likely unreliable. The other data sets do not seem to contain such values, at least based on the first sight. All data sets contain missing values, what is hard to notice in Figure~\ref{fig:initial_plots} due to the high frequency of the data.

\begin{figure}[b!]
	\includegraphics[width=1\linewidth]{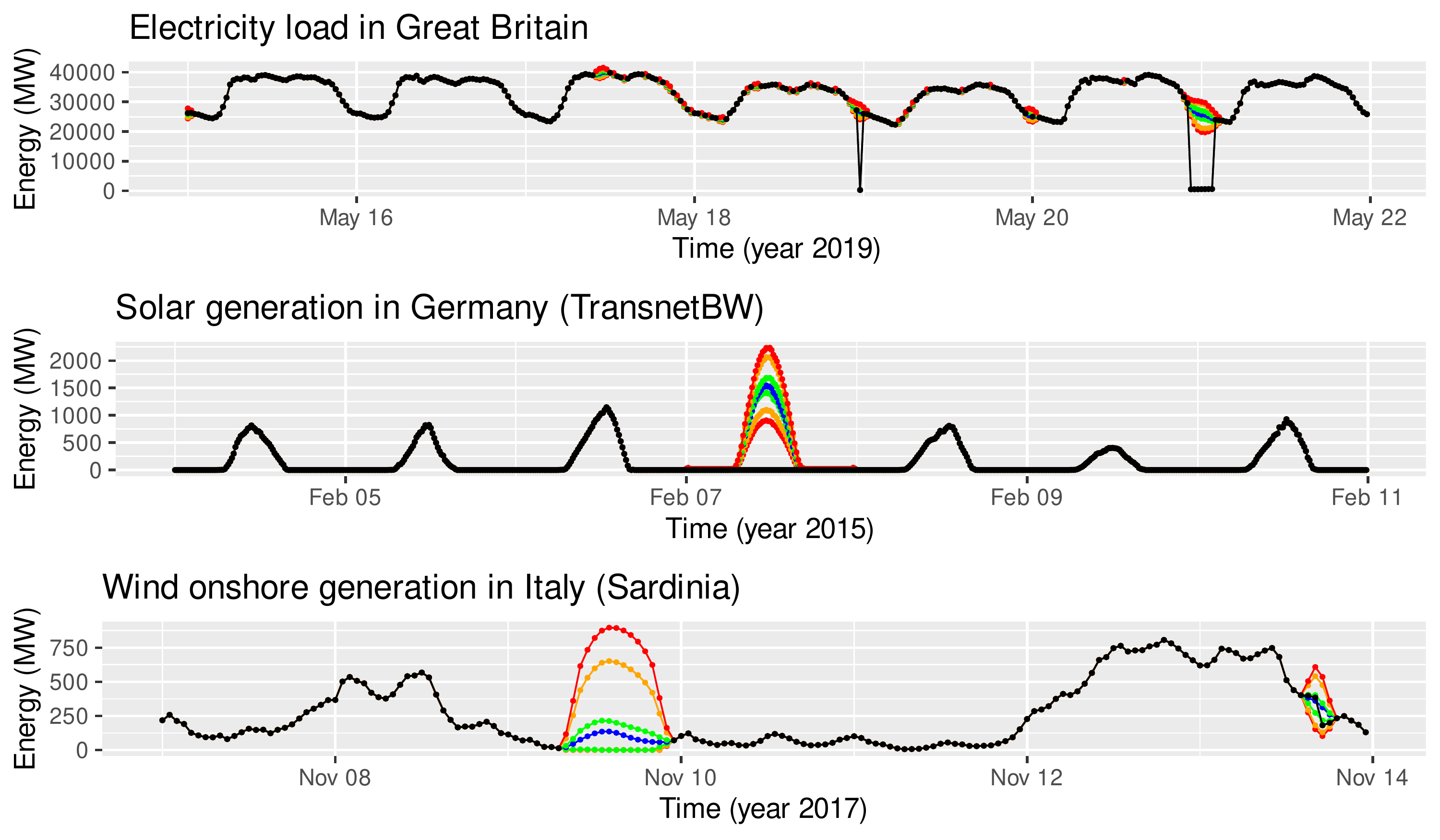}
	
	\caption{Chosen parts of the data that illustrate the problem with replaced missing values and unreliable outliers. Red lines indicate 95\% confidence intervals, orange 90\%, green 50\%, and blue indicates the median.}
	\label{fig:chosen_dates}
\end{figure}

For a better illustration, a subset of each series containing either missing values, outliers or both is depicted in Figure~\ref{fig:chosen_dates}. The time series of electricity load in Great Britain exhibits both missing values and unreliable outliers. The latter ones are quite easy to spot by eyeballing --- they are the two negative spikes visible in the plot. 
The time series of solar generation in Germany contains unreliable outliers, however, much harder to detect. Namely, all the observations of February 07, 2015 are set to zero, what is impossible even for a very cloudy day. Such outliers are especially challenging as half of the observations are equal to zero. The third example, the wind onshore generation in Italy, contains no unreliable values but a clear gap in the data on November 09, 2017. 

\begin{figure}[b!]
	\includegraphics[width=1\linewidth]{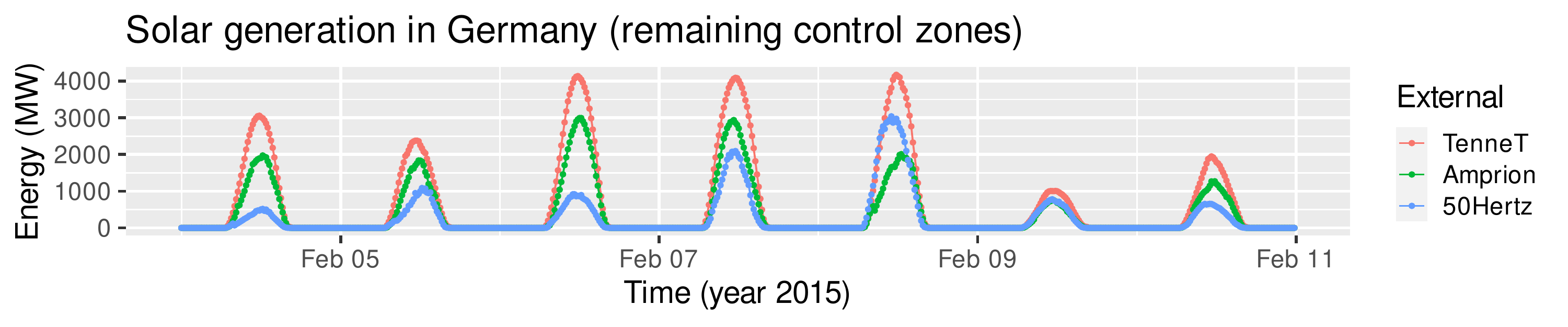}
	
	\caption{The solar generation in Germany in remaining control zones used as external regressors in our exercise.}
	\label{fig:solar_externals}
\end{figure}

The cleansing of the named series was done with the \verb+auto_data_cleaning+ function. The first time series, electricity load in Great Britain, and the third time series, wind onshore generation in Italy, were handled in a univariate setting without external inputs with
$\boldsymbol{S} = (48, 7\cdot48)$ and $\boldsymbol{S} = 24$, respectively. To account for the unreliable zeros during the day shown in Figure~\ref{fig:chosen_dates} the \verb+consider.as.missings+ parameter has been used to recognize zeros as missing values only if the whole day consists of zeros. The second series, the solar generation in Germany, has been handled with $\boldsymbol{S} = 96$ in a multivariate setting such that external regressors as plotted in Figure~\ref{fig:solar_externals} were utilized for data modelling and imputation. Additionally, we set the \verb+whole.period.missing.only+ parameter to \verb+TRUE+ to consider zeros as missings only if the whole day is zero, and we set \verb+replace.recursively+ to \verb+FALSE+ to avoid error accumulation during night hours. All the other arguments were set to their default values. The models were estimated for quantiles $\tau \in \{0.025, 0.05, 0.25, 0.5, 0.75, 0.95, 0.975\}$.

The results provided in Figure~\ref{fig:chosen_dates} are very promising. This accounts in particular for the first data set, where only autoregressive effects were used. It can be seen that the outliers were smoothly replaced, i.e. not only the outliers themselves were replaced, but also their direct neighbours. The cleaning of the other two data sets is also satisfying and the values seem to be reasonable. 

\begin{figure}[b!]
	\includegraphics[width=1\linewidth]{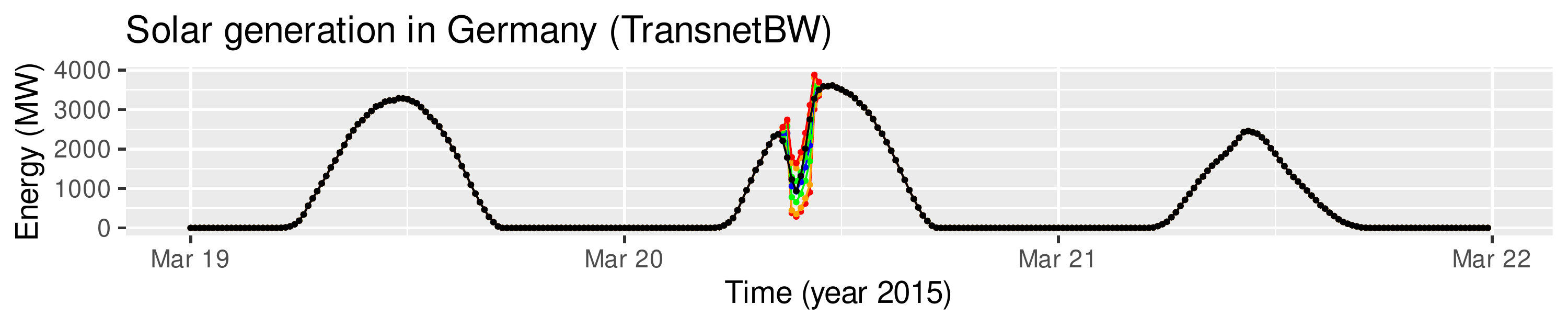}
	
	\caption{A drop in the solar generation in Germany that may seem to be unreliable while it was caused by a natural phenomenon, i.e. the partial solar eclipse. Red lines indicate 95\% confidence intervals, orange 90\%, green 50\%, and blue indicates the median.}
	\label{fig:solar_eclipse}
\end{figure}

Although the algorithm works well and robust, it has to be noted that specific causes of an anomaly (e.g. technical failures, true behavioural issues or nature phenomenons) are not always clear distinguishable and detectable. An example of such situation is presented in Figure~\ref{fig:solar_eclipse}. Here, the algorithm is able to detect the anomalous negative spike. However, due to the fact that this spike corresponds to a partial solar eclipse, the detection is considered as false positive. By providing external inputs, i.e. generation in the other German control zones, the algorithm is able to correct this misclassification by imputing very similar values. However, this example shows us that even though the provided package works robust, a supervision by humans, in particular when dealing with outliers and extreme values, is recommended.

\section{Performance evaluation against competitors}
To verify the potential of the proposed package, its functions are compared and evaluated against benchmarks from the literature. The performance of \verb+model_missing_data+ is evaluated in an imputation study.
Here, the electricity load in Great Britain ranging form January 01, 2019 to December 31, 2019 is used. To allow a performance evaluation the named data set is manipulated by replacing existing values by missing entries. Observations are removed in continuous blocks. Assuming that values are missing at random (MAR) such that the likelihood of an observation missing depends on whether observations closer in time have also been missing. This sampling approach is implemented by a log-normal distribution parametrized with $\mu = 12$ and $\sigma = 6$. To account for a varying number of missing values, a share of 0.01, 0.05, 0.1, 0.2 and 0.5 missing values is considered. Each simulation is repeated 20 times to allow a robust evaluation. As most methods in the literature are restricted to univariate time series data, the imputation study is implemented in a univariate setting. 
As competitors the \verb+impute_AR1_t+ function of the \verb+imputeFin+ package \cite{Liu2021}, the \verb+na.StructTS+, \verb+na.locf+ and \verb+na.aggregate+ functions of the \verb+zoo+ package \cite{Zeileis2005}, the \verb+na_ma+, \verb+na.kalman+ and \verb+na_interpolation+ functions of the \verb+imputeTS+ package \cite{Moritz2017}, the \verb+prophet+ function of the \verb+prophet+ package \cite{Taylor2020} as well as the \verb+ts_impute_vec+ of the \verb+timetk+ \cite{timetk} package are applied.  All methods are provided with default settings and named in accordance to the package and function name.
As evaluation measures the mean absolute error (MAE) and the computational time (single-core on Intel Core i5-8250U CPU 1.80GHz) are used. The results are shown in Table~\ref{tab:Imputation_study}. It can be seen that the proposed function \verb+model_missing_data+ dominates all considered benchmarks for each simulation in terms of the MAE. In terms of the computational time, the \verb+model_missing_data+ function performs moderate in comparison to the benchmarks. \

\begin{table}[b!]
	\centering
	\begingroup\small
	\begin{adjustbox}{max width=1\textwidth}
	\begin{tabular}{rrrrrr}
  \hline
 & 0.01 & 0.05 & 0.1 & 0.2 & 0.5 \\ 
  \hline
\verb+tsrobprep::model_missing_data+ & \cellcolor[rgb]{0.5,0.9,0.5} {\textbf{623 (4.634)}} & \cellcolor[rgb]{0.5,0.9,0.5} {\textbf{603 (8.907)}} & \cellcolor[rgb]{0.5,0.9,0.5} {\textbf{631 (14.837)}} & \cellcolor[rgb]{0.5,0.9,0.5} {\textbf{657 (23.302)}} & \cellcolor[rgb]{0.5,0.9,0.5} {\textbf{828 (43.821)}} \\ 
  \verb+imputeTS::na_seadec.interpolation+ & \cellcolor[rgb]{0.73,0.977,0.5} {926 (0.062)} & \cellcolor[rgb]{0.679,0.96,0.5} {849 (0.063)} & \cellcolor[rgb]{0.676,0.959,0.5} {880 (0.063)} & \cellcolor[rgb]{0.697,0.966,0.5} {932 (0.06)} & \cellcolor[rgb]{0.84,1,0.5} {1317 (0.06)} \\ 
  \verb+prophet::prophet+ & \cellcolor[rgb]{1,0.948,0.5} {1822 (58.115)} & \cellcolor[rgb]{1,0.958,0.5} {1779 (61.191)} & \cellcolor[rgb]{1,0.957,0.5} {1849 (56.287)} & \cellcolor[rgb]{1,0.961,0.5} {1828 (53.033)} & \cellcolor[rgb]{1,0.98,0.5} {1809 (61.058)} \\ 
  \verb+imputeTS::na_kalman.auto.arima+ & \cellcolor[rgb]{1,0.94,0.5} {1887 (33.148)} & \cellcolor[rgb]{1,0.917,0.5} {2112 (27.029)} & \cellcolor[rgb]{1,0.886,0.5} {2451 (18.33)} & \cellcolor[rgb]{1,0.882,0.5} {2488 (11.76)} & \cellcolor[rgb]{1,0.804,0.5} {3235 (9.526)} \\ 
  \verb+imputeTS::na_interpolation.stine+ & \cellcolor[rgb]{1,0.895,0.5} {2242 (0.005)} & \cellcolor[rgb]{1,0.865,0.5} {2542 (0.006)} & \cellcolor[rgb]{1,0.855,0.5} {2720 (0.006)} & \cellcolor[rgb]{1,0.842,0.5} {2828 (0.006)} & \cellcolor[rgb]{1,0.763,0.5} {3574 (0.006)} \\ 
  \verb+timetk::ts_impute_vec+ & \cellcolor[rgb]{1,0.831,0.5} {2749 (1.525)} & \cellcolor[rgb]{1,0.808,0.5} {3016 (1.567)} & \cellcolor[rgb]{1,0.797,0.5} {3209 (1.492)} & \cellcolor[rgb]{1,0.783,0.5} {3323 (1.397)} & \cellcolor[rgb]{1,0.71,0.5} {4003 (1.589)} \\ 
  \verb+imputeTS::na_ma.exponential+ & \cellcolor[rgb]{1,0.766,0.5} {3263 (0.002)} & \cellcolor[rgb]{1,0.754,0.5} {3460 (0.005)} & \cellcolor[rgb]{1,0.745,0.5} {3653 (0.01)} & \cellcolor[rgb]{1,0.735,0.5} {3721 (0.02)} & \cellcolor[rgb]{1,0.666,0.5} {4357 (0.058)} \\ 
  \verb+imputeFin::impute_AR1_t+ & \cellcolor[rgb]{1,0.746,0.5} {3417 (13.768)} & \cellcolor[rgb]{1,0.721,0.5} {3730 (16.703)} & \cellcolor[rgb]{1,0.723,0.5} {3838 (19.709)} & \cellcolor[rgb]{1,0.702,0.5} {4000 (19.611)} & \cellcolor[rgb]{1,0.629,0.5} {4658 (27.522)} \\ 
  \verb+zoo::na.locf+ & \cellcolor[rgb]{1,0.561,0.5} {4881 (0.001)} & \cellcolor[rgb]{1,0.533,0.5} {5285 (0.001)} & \cellcolor[rgb]{1,0.555,0.5} {5268 (0.003)} & \cellcolor[rgb]{1,0.519,0.5} {5537 (0.001)} & \cellcolor[rgb]{1,0.5,0.535} {6000 (0.002)} \\ 
  \verb+zoo::na.aggregate.median+ & \cellcolor[rgb]{1,0.5,0.55} {5758 (0.002)} & \cellcolor[rgb]{1,0.5,0.55} {5968 (0.002)} & \cellcolor[rgb]{1,0.5,0.55} {6163 (0.006)} & \cellcolor[rgb]{1,0.5,0.55} {6120 (0.002)} & \cellcolor[rgb]{1,0.5,0.55} {6119 (0.004)} \\ 
   \hline
\end{tabular}
\end{adjustbox}
\caption{Results of the imputation study for varying shares of missing values in terms of the MAE in MW (and the computational time in seconds). Colour indicates the performance column-wise (the greener, the better). With bold, we depicted the best values in each column.}
	\label{tab:Imputation_study}
	\endgroup
\end{table}

\begin{figure}[t!]
	
	\includegraphics[width=1\linewidth]{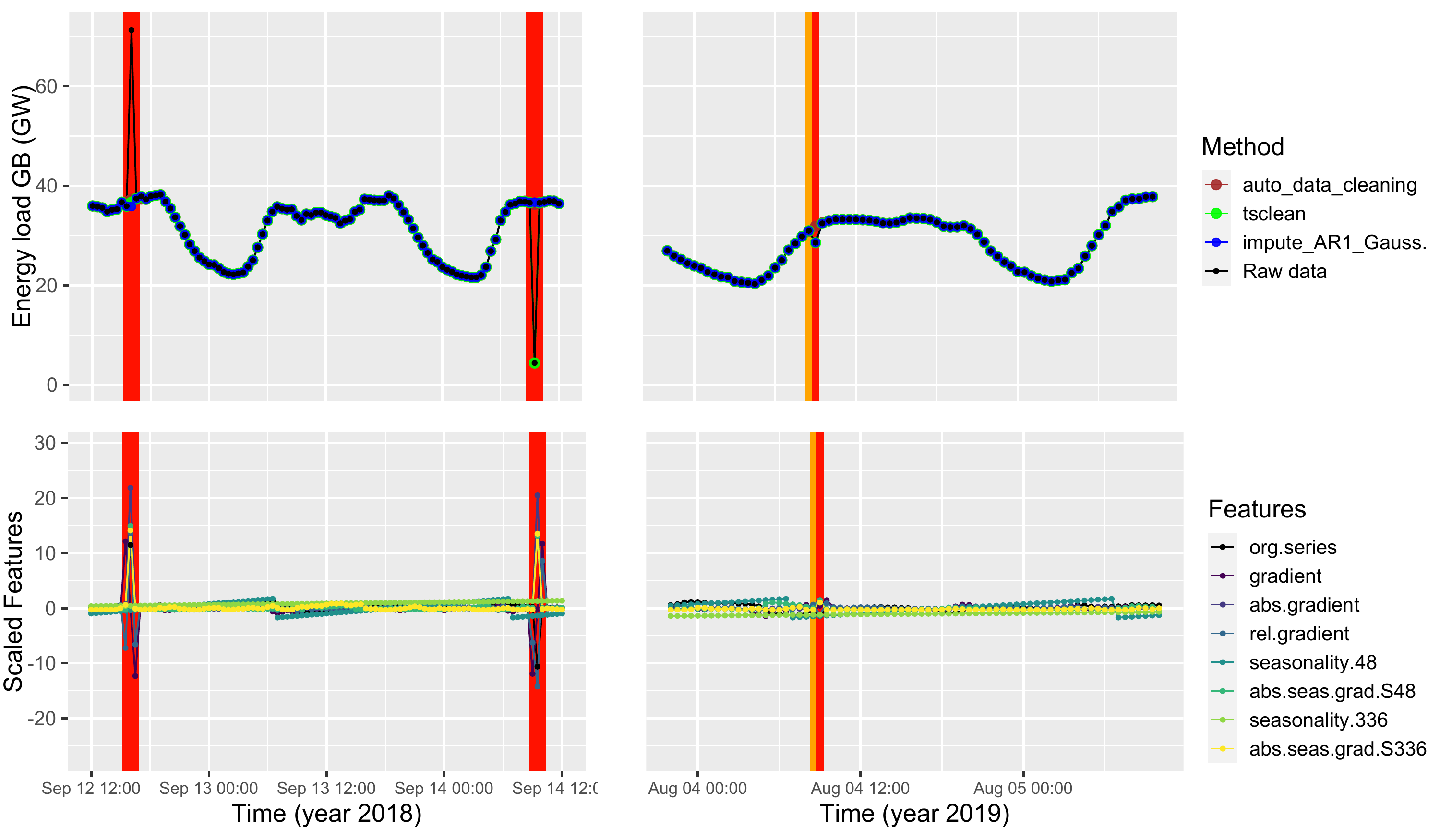}
	
	\caption{Exemplary comparison of the proposed function with competitors from literature on the electricity load in Great Britain. The upper plots display the cleansed data of the considered methods and the lower plots the feature space used for outlier detection of the detect\_outliers function. The coloured vertical lines indicate the probability of a specific observation being outlying data, ranging from dark red (high probability) to yellow (low probability). Probabilities below 0.001 are not coloured.}
	\label{fig:auto_data_cleaning_example}
\end{figure}

Focusing on the second core function \verb+detect_outliers+ and the wrapper function \verb+auto_data_cleaning+, a confident evaluation is not straight-forward as the definition of outliers can not be easily narrowed down in a generally valid way. Hence, the performance is presented illustratively by comparing the cleansed data set returned by the \verb+auto_data_cleaning+ function with the results of two benchmark procedures from the literature, namely the \verb+tsclean+ function of the \verb+forecast+ package \cite{Hyndman2008} and the \verb+impute_AR1_t+ function of the \verb+imputeFin+ package \cite{Liu2021}. As data set the electricity load in Great Britain without any manipulation ranging from January 01, 2018 to December 31, 2019 is used. Two examples are shown in Figure~\ref{fig:auto_data_cleaning_example}. It is observable that the proposed procedure \verb+auto_data_cleaning+ detects all three anomalies in the data. The outlying instances on the left side of the Figure are clearly technical failures and the outlying data on the right side of the Figure is at least questionable. By considering the lower plots of Figure~\ref{fig:auto_data_cleaning_example}, the feature space of the \verb+detect_outliers+ function is displayed. The vertical lines indicate the probability of each observation being outlying data.  
The results provided in Table~\ref{tab:Imputation_study} and Figure~\ref{fig:auto_data_cleaning_example} are promising and underline the suitability, especially, for energy system data.

\section{Impact and conclusions}
So far the available R packages for time series data preprocessing are based on either simple and not robust or on complex and very slow algorithms. The introduction of the tsrobprep package fills this gap by providing a robust and efficient method to replace missing values and handle unreliable outliers. Moreover, by providing the function \verb+auto_data_cleaning+, the package is simple applicable and provides already with the default parameter setting satisfying results. More advanced users can take advantage of the big set of tuning parameters to further adapt the functions to their needs.

The software has been already used in the purpose of academic research~\cite{Narajewski2020} and industrial projects in the area of power plant electricity generation. Moreover, the software is written in such a manner that it can be utilized on a daily basis with almost no effort. The package is open-source, available online in the CRAN repository, well-documented and constantly developed. The presented examples have shown that the package might be a powerful tool in various time series data preprocessing exercises.

\section{Conflict of Interest}

No conflict of interest exists:
We wish to confirm that there are no known conflicts of interest associated with this publication and there has been no significant financial support for this work that could have influenced its outcome.



\bibliographystyle{elsarticle-num} 


\begin{table}[!h]
	\begin{tabular}{|l|p{6.5cm}|p{6.5cm}|}
		\hline
		\textbf{Nr.} & \textbf{(Executable) software metadata description} & \textbf{Please fill in this column} \\
		\hline
		S1 & Current software version & 0.3.1 \\
		\hline
		S2 & Permanent link to executables of this version  & \url{https://cran.r-project.org/web/packages/tsrobprep} \\
		\hline
		S3 & Legal Software License &  MIT license (MIT)  \\
		\hline
		S4 & Computing platforms/Operating Systems & Linux, OS X, Microsoft Windows\\
		\hline
		S5 & Installation requirements \& dependencies &  R ($\ge$ 3.2.0) \\
		\hline
		S6 & If available, link to user manual - if formally published include a reference to the publication in the reference list &  \url{https://cran.r-project.org/web/packages/tsrobprep/tsrobprep.pdf} \\
		\hline
		S7 & Support email for questions & michal.narajewski@uni-due.de \\
		\hline
	\end{tabular}
\end{table}

\end{document}